\documentclass[conference,compsoc]{IEEEtran}
\usepackage{xcolor}
\usepackage{amsmath}
\usepackage{amssymb}
\usepackage{enumitem}
\usepackage{colortbl}
\definecolor{mygray}{gray}{0.8}
\usepackage{flushend}
\usepackage{balance}
\usepackage{booktabs}
\usepackage{graphicx}
\usepackage{amsmath}
\usepackage[ruled,vlined]{algorithm2e}

\usepackage{xcolor}
\definecolor{deeppurple}{rgb}{0.40, 0.20, 0.60}
\usepackage{lipsum}
\usepackage{subfig}
\usepackage{color}

\usepackage[utf8]{inputenc} 
\usepackage{bigstrut}
\usepackage{multirow}
\usepackage{siunitx}
\usepackage{listings}
\usepackage{eso-pic}
\usepackage{tikz}
\usepackage{xspace}
\usepackage{fancyhdr}
\usepackage{mathtools}
\usepackage{soul}
\usepackage[leftcaption]{sidecap}
\usepackage{diagbox}
\usepackage{makecell}
\usepackage{hyperref}

\ifCLASSOPTIONcompsoc
  \usepackage[nocompress]{cite}
\else
  \usepackage{cite}
\fi
\ifCLASSINFOpdf
\else
\fi
\newcommand{\fixme}[1]{%
  \ifmmode
    \textcolor{blue}{#1}
  \else
    \textcolor{blue}{\textit{#1}}
  \fi
}

\newcommand{\fixdel}[1]{%
  \ifmmode
    \textcolor{red}{\cancel{#1}}%
  \else
    \textcolor{red}{\sout{#1}}%
  \fi
}

\ifx\figurename\undefined \def\figurename{Figure}\fi
\renewcommand{\figurename}{Fig.}
\renewcommand{\paragraph}[1]{\textbf{#1} }

\newcommand{\Sect}[1]{Sec.~\ref{#1}}
\newcommand{\Fig}[1]{Fig.~\ref{#1}}
\newcommand{\Tbl}[1]{Tbl.~\ref{#1}}
\newcommand{\Equ}[1]{Equ.~\ref{#1}}

\newcommand{\Alg}[1]{Algo.~\ref{#1}}

\newcommand{\proj}{\textsc{Annie}\xspace}

\newcommand{\no}[1]{#1}
\renewcommand{\no}[1]{}
\newcommand{\RNum}[1]{\uppercase\expandafter{\romannumeral #1\relax}}

\begin{document}
%
\title{\proj: Be Careful of Your Robots}



%
\author{\IEEEauthorblockN{
Yiyang Huang\IEEEauthorrefmark{1},
Zixuan Wang\IEEEauthorrefmark{2},
Zishen Wan\IEEEauthorrefmark{3}\IEEEauthorrefmark{5}, 
Yapeng Tian\IEEEauthorrefmark{4},
Haobo Xu\IEEEauthorrefmark{1},
Yinhe Han\IEEEauthorrefmark{1},
and
Yiming Gan\IEEEauthorrefmark{1}\IEEEauthorrefmark{5}}
\IEEEauthorblockA{\IEEEauthorrefmark{1}Institute of Computing Technology, Chinese Academy of Sciences, Beijing, China}
\IEEEauthorblockA{\IEEEauthorrefmark{2}Institute of Automation, Chinese Academy of Sciences, Beijing, China}
\IEEEauthorblockA{\IEEEauthorrefmark{3}Georgia Institute of Technology, Atlanta, GA, USA}
\IEEEauthorblockA{\IEEEauthorrefmark{4}University of Texas at Dallas, Richardson, TX, USA}
\IEEEauthorblockA{\IEEEauthorrefmark{5}Corresponding authors}
}


\maketitle

\begin{abstract}



The integration of vision-language-action (VLA) models into embodied AI (EAI) robots is rapidly advancing their ability to perform complex, long-horizon tasks in human-centric environments. However, EAI systems introduce critical \emph{security risks}: a compromised VLA model can directly translate adversarial perturbations on sensory input into unsafe physical actions. Traditional safety definitions and methodologies from the machine learning community are no longer sufficient. EAI systems raise new questions, such as what constitutes safety, how to measure it, and how to design effective attack and defense mechanisms in physically grounded, interactive settings.

In this work, we present the \emph{first systematic study of adversarial safety attacks on embodied AI systems}, grounded in ISO standards for human-robot interactions. We (1) formalize a principled taxonomy of safety violations (critical, dangerous, risky) based on physical constraints such as separation distance, velocity, and collision boundaries; (2) introduce \emph{\proj-Bench}, a benchmark of nine safety-critical scenarios with 2,400 video-action sequences for evaluating embodied safety; and (3) \emph{\proj-Attack}, a task-aware adversarial framework with an attack leader model that decomposes long-horizon goals into frame-level perturbations.

Our evaluation across representative EAI models shows attack success rates exceeding 50\% across all safety categories. We further demonstrate \emph{sparse and adaptive attack strategies} and validate the real-world impact through physical robot experiments. These results expose a previously underexplored but highly consequential attack surface in embodied AI systems, highlighting the urgent need for security-driven defenses in the physical AI era. Code is available at https://github.com/RLC-Lab/Annie.

\end{abstract}
\section{Introduction}
We are entering the era of physical artificial intelligence (AI). In contrast to semantic AI, which primarily operates in abstract or virtual domains, physical AI integrates deep learning models into real-world environments, enabling AI agents to interact with their surroundings and to receive feedback. This shift marks a significant step toward achieving artificial general intelligence (AGI) by grounding intelligence in perception, action, and continuous adaptation. Embodied AI (EAI) systems, which equip robots with advanced AI capabilities, represent one of the most prominent and representative applications of physical AI and are now emerging in the fields~\cite{kimopenvla,haldarbaku,huang2024corki,wan2025reca,mu2023embodiedgpt,mandi2024roco,sun2024dadu}.

Traditionally, robot safety has been a well-studied and relatively mature field~\cite{rubagotti2022perceived,bi2021safety}. In industrial settings, where robots operate with high-speed, high-power actuators, humans are typically restricted from entering the workspace to prevent physical harm. For robots designed to operate in human living spaces, safety has historically been ensured by three key factors. First, such robots are typically equipped with low-speed, low-power motors, which inherently limit the force and velocity of their movements. Second, their behaviors are explicitly programmed by experienced engineers, with motion paths carefully designed to avoid any unintended contact with humans. Third, the tasks assigned to these robots are typically simple and routine, involving limited use of tools and minimal variation. As a result, the robots operate with low autonomy, further reducing the risk of unexpected or unsafe behavior. Things change significantly in EAI systems, particularly with respect to the second and third factors. Robots are now being developed to handle complex instructions and interact with environments~\cite{zhang2024building,zhang2024combo,wu2023embodied,wan2024thinking,wan2025generative}.


Such a trend leads to new safety concerns~\cite{zhang2024safeembodai,xing2025towards}, and we identify three fundamental questions that remain unresolved. First, the definition of safety in EAI systems is still unclear. Many existing studies conflate safety hazards with general performance degradation~\cite{zhou2025badvla,cheng2024manipulation}, treating any attack that causes malfunction or failure as a safety issue. However, an attack that merely results in task failure or a robot stall may not constitute a safety violation. The prerequisite for studying safety in EAI systems is to establish a clear and principled definition of what constitutes safe or unsafe behavior.

Second, there is a lack of dedicated datasets for safety research. Although the EAI domain has produced numerous rich datasets~\cite{gan2021threedworld,puig2021watch,lin2023mcu}, these primarily aim to improve model generalizability and task performance across diverse scenarios. In contrast, safety-centric datasets should focus on corner cases—rare or high-risk situations where robot behavior may pose safety concerns. Such datasets should include curated scenarios designed to trigger potential safety violations, along with training and evaluation sequences that allow researchers to test the effectiveness of attack and defense methodologies. In addition, new safety-specific metrics are needed to appropriately measure model performance within the context of safety-critical outcomes.

The third unresolved question is the lack of effective attack and defense strategies for EAI systems, especially for the action chains that directly interact with physical environments through vision-to-action models. While both attack and defense methodologies have been proposed for EAI, the absence of clear safety definitions and metrics has led many studies to simply adopt conventional approaches from adversarial research in deep learning~\cite{liu2020spatiotemporal,chen2024towards}. For instance, most existing attacks still focus on optimizing perturbations to minimize pixel-level differences while maximizing performance degradation—an approach inherited from image classification tasks. However, EAI systems introduce a fundamentally different context, involving long-horizon video, physical embodiment with real-world interaction, and dynamic feedback, which opens up a vast and largely unexplored space for developing safety-specific attack and defense strategies.

In this work, we endeavor to be the \emph{first} to address the three unresolved questions outlined above. For the safety definition, we move beyond traditional machine learning practices that rely on accuracy-based metrics. Instead, we draw from ISO/TS 15066, a widely recognized standard in human–robot collaboration, to define what constitutes a safety violation. Building on this foundation, we propose a classification scheme that categorizes safety violations in EAI applications into three levels: \textit{critical}, \textit{dangerous}, and \textit{risky}, based on their potential impact on human safety.

Building on our safety definition for EAI systems, we construct the first safety-centric benchmark. The dataset includes 9 in-house scenarios, each representing a task with potential safety violation corner cases. From these scenarios, we collect over 2,400 video-action sequences for training and testing, covering all three categories of safety violations: critical, dangerous, and risky. To support future research on attacks and defenses, we also introduce a set of evaluation metrics that assess whether a given action sequence results in a safety violation, as well as the consistency and deviation of the action execution. 


Finally, leveraging our dataset, we propose a novel attack that specifically targets the most critical component in the action chain of EAI robots: the vision-language-action (VLA) model. Our attack framework addresses the unique challenge of performing adversarial attacks on long video-action sequences, where the precise optimization goal during the attack is often missing or unclear. We introduce a novel attack leader model that translates the long-term goal into separate frame-based attack targets.  

We evaluate our attack framework on two representative VLA models, ACT~\cite{zhao2023learningfinegrainedbimanualmanipulation} and Baku~\cite{haldarbaku}. The attack framework successfully results in safety violations at all three levels. The average attack success rate is 52\% on critical applications, 67\% on dangerous applications, and 50\% on risky applications. The attack is conducted smoothly and hiddenly, with a very high consistency between previous actions and following actions.

To summarize, this paper has the following contributions:
\begin{itemize}
    \item We introduce principled definitions and metrics of EAI system safety, which are new to this area. To ground this definition, we draw from established human-robot collaboration standards, specifically the safety regulations outlined in ISO/TS 15066.
    \item We propose a benchmark focused on EAI safety, comprising 9 different scenarios that cover all categories of unsafe behavior that violate our definition. The benchmark includes 2,400 vision-action sequences, providing a comprehensive resource for evaluating safety-critical performance in embodied AI systems.
    \item We introduce an attack framework targeting EAI systems, achieving an attack success rate exceeding 50\% across all safety-related scenarios in our benchmark.
\end{itemize}

\section{Background}
\label{sec:bg}
We introduce the safety concerns in human-robot collaboration (\Sect{sec:bg:hrcsafe}), the concept of EAI systems (\Sect{sec:bg:eai}), the concept of adversarial attacks (\Sect{sec:bg:advatt}), and safety concerns in EAI systems (\Sect{sec:bg:eaisafe}). 


\subsection{Human-robot Collaboration Safety}
\label{sec:bg:hrcsafe}
Since robots were first introduced into human living spaces, safety standards for human-robot collaboration have become critically important. Strict regulations are applied to robot structures, actuators, and control algorithms to ensure safety. Thus, International organizations like the International Organization for Standardization (ISO) and the International Electrotechnical Commission (IEC) have established relevant global standards\cite{li2024safe} on robots and human-robot collaboration (HRC) systems.


These standards are categorized into three types: Type A standards provide the foundational principles for machinery design; Type B standards address general aspects applicable to a wide range of machines; and Type C standards focus on safety requirements for specific types of machinery\cite{VILLANI2018248}.

Among above three standards, Type A and B are primarily designed for traditional industrial robots, mainly setting regulations to the mechanical components of a robot system, including its structure and speed, such as ISO 13849\cite{iso13849} and ISO 13855\cite{ISO13855-2024} standards in Type B. Tybe C, on the other hand, differs from the above two types and targets on human robot collaborations. Type C standard ISO/TS 15066\cite{ISO15066} is the most relevant and authoritative international guidelines explicitly tailored for collaborative robots\cite{li2024safe}. ISO/TS 15066 outlines safety-related recommendations, including four primary modes of collaborative operation:

\begin{itemize}
    \item \textbf{Safety-Rated Monitored Stop (SRMS)}: In this mode, the robot and human cannot work in the shared space at the same time. The robot operates until a human enters its workspace, which triggers an automatic stop via safety sensors (e.g., cameras, LiDAR, light curtains), and only resumes operation after the human leaves.
    \item \textbf{Hand Guiding (HG)}: Building on SRMS, this mode allows the human to manually guide the robot using a dedicated device, enabling safe, direct interaction within a controlled range.
    \item \textbf{Speed and Separation Monitoring (SSM)}: The robot and human can operate simultaneously, provided a safe distance is maintained. Sensors monitor this distance in real time, and the robot's speed adjusts dynamically based on human proximity, following ISO 13855 and ISO/TS 15066.
    \item \textbf{Power and Force Limiting (PFL)}:  This mode permits close physical interaction by ensuring the robot's contact force stays within safe limits. Safety is achieved through both passive design elements (e.g., soft surfaces, rounded edges) and active control strategies.
\end{itemize}

We find that the above standards are suitable for EAI robots, as they are heavily evolved in human-robot collaboration tasks. Hence, in our work, the definition of safety is primarily informed by the guidelines set forth in ISO/TS 15066 standard. 



\subsection{EAI Systems}
\label{sec:bg:eai}
Traditionally, robots have operated using rule-based systems. To complete a given task, experienced programmers must explicitly define the starting position, ending position, and even the exact trajectory, resulting in limited autonomy.

With the advancement of deep learning algorithms, particularly the emergence of large language models (LLMs), the landscape of robotics has changed. EAI, which refers to artificial intelligence systems that interact with their environment through a physical robotic body, is becoming a dominant trend~\cite{xu2024survey}. EAI systems enable robots to perform complex tasks in open environments without guidance or assistance from human engineers. Specifically, they enhance two critical components of the robotic computing pipeline: the reasoning chain and the action chain.

\paragraph{Reasoning Chain.} The reasoning chain in EAI systems refers to the cognitive process that guides robots to finish long-horizon tasks. It involves interpreting high-level goals, understanding the environment, decomposing tasks into actionable steps, and adapting to dynamic changes during execution. This chain enables robots to operate autonomously in unstructured settings by making informed decisions based on sensory input, task objectives, and contextual constraints. The reasoning chain is dominated by diverse LLMs~\cite{song2023llm,huang2023voxposer,rana2023sayplan,han2024llm,wang2024karma,sun2024dadu}.


\paragraph{Action Chain.} The action chain in embodied AI robots refers to the execution process that transforms planned decisions into physical movements. It includes motion planning, trajectory generation, control, and interaction with objects or humans in the environment. The action chain is equally important as the reasoning chain, but it covers completely different aspects of embodied AI robots~\cite{zawalski2024robotic,zhao2025cot}. Unlike LLMs in the reasoning chain, vision-language models (VLMs) and vision-language-action (VLA) models play an increasingly important role in strengthening the action chain~\cite{brohan2022rt,brohan2023rt,fu2024mobile,team2024octo,cheang2024gr,kim2024openvla,haldarbaku}. 



\subsection{Adversarial Attacks}
\label{sec:bg:advatt}

EAI systems are centered around deep learning algorithms, which makes them particularly susceptible to input perturbations. Adversarial attacks exploit this weakness by introducing carefully crafted, minimal perturbations to the input. The objective is to compromise embodied AI systems-a robot-by deceiving deep learning models utilized in perception and decision-making processes.



Formally, in EAI, the adversarial attack objective can be expressed in \Equ{equ:advatt_embodiment}:

\begin{equation}
    \label{equ:advatt_embodiment}
   \arg \min _ { \theta } \| \theta \| ~~ \text { s.t. } ~~ \Phi ( f ( X + \theta ) ) \notin \mathcal{S} _ { \text { safe } } ,
\end{equation}

Where $X$ is the original sensory input, $f$ denotes the EAI policy, and $\theta$ represents adversarial perturbations.  Here, $\phi$ denotes the resulting state after executing the action, which encompasses the status of objects, humans, the robot itself, and so on within the environment. An attack is considered successful when the executed action outcome $ \Phi ( f ( X + \theta ) )$ violates the predefined safety state set $ \mathcal{S} _ { \text { safe } } $, thereby driving the agent into unsafe or undesirable states.

Studies have demonstrated the effectiveness of adversarial attacks~\cite{goodfellow2014explaining,carlini2017towards,yuan2019adversarial}. Based on the attack model, these attacks are generally classified into two categories: white-box attacks, where the attacker has full access to the target model's architecture, parameters, and gradients; and black-box attacks, where the attacker has limited knowledge and can only observe the model's inputs and outputs.


\paragraph{Defense.} Researchers have proposed various defense methods~\cite{madry2017towards,meng2017magnet,xu2017feature}. Based on when the defense is applied, these methods can be broadly classified into training-time defenses, such as adding adversarial examples into the training set~\cite{xie2020adversarial,han2023interpreting}, and testing-time defenses including input randomization and transformation~\cite{meng2017magnet,xu2017feature,xie2017mitigating,guo2017countering}.



\subsection{Safety Concerns in EAI Systems}
\label{sec:bg:eaisafe}
EAI systems can be broadly divided into reasoning chains and action chains, both of which raise important safety concerns. While this paper \textbf{focuses primarily on the safety of the action chain}, we introduce both components in this section to provide a complete view of the system architecture and its associated risks.

\subsubsection{Safety in the Reasoning Chain}
The core of the reasoning chain in EAI systems is typically an LLM. Despite significant efforts to prevent LLMs from generating harmful content~\cite{wen2024secure,liu2024exploring}, studies have shown that they remain vulnerable to carefully crafted adversarial prompts designed to elicit unsafe or unintended behavior.

One of the most prominent examples is the jailbreak attack, where carefully crafted prompts often combined with iterative question-and-answer interactions can induce an LLM to generate outputs that violate established safety standards~\cite{pathade2025red,lu2024poex}. In the context of EAI, such attacks may lead the model to produce unsafe instructions for robots. For instance, during a household task, a compromised LLM might generate a command such as pouring hot water on a human, which poses a clear safety hazard~\cite{liu2024exploring, yin2024safeagentbench, lu2024poex, zhang2024badrobot}.


\subsubsection{Safety in the Action Chain}

Safety concerns in the action chain involve a different attack model. Unlike jailbreak attacks, which induce LLMs to generate harmful instructions, attacks on the action chain aim to alter the robot's physical behavior despite receiving benign instructions. These attacks target the perception, planning, or control modules to produce unsafe actions, potentially leading to unintended interactions with the environment or humans.

With the increasing adoption of end-to-end trained VLA models in the action chains of EAI systems, these models have become the primary targets of attack. While attacks on VLA models and traditional convolutional neural networks (CNNs) often share the same input modality (e.g., images), there are two key differences that distinguish them.

First, attacks on CNNs operate on a single image, with their effects observed immediately in the model's output. In contrast, attacks on VLA models are often applied to a sequence of images or a video, as these models process temporal information to generate actions. Second, the success criteria for attacks differ significantly. In CNN-based tasks, an attack is typically considered successful if it degrades prediction accuracy, for example, by causing misclassification or incorrect bounding box predictions in detection tasks. However, in VLA models, simply reducing task success rates may not be a meaningful measure.



\section{EAI Safety Definition and Classification}
\label{sec:def}

\subsection{Safety Definition}
\label{sec:def:def}

VLAs are now the dominant approach in the EAI action chain. A VLA typically takes in a set of observation $o_t$, which are mainly images $I _ { t } ^ { 1 }, \ldots, I _ { t } ^ { n }$,  but can be with various sensor inputs in different VLA models\cite{intelligence2025pi05visionlanguageactionmodelopenworld,yan2024robomm,haldarbaku}, and a natural language task instruction $l$. 


\begin{equation}
\label{equ:vla}
    a_t = VLA(o_t,l)
\end{equation}

The output is the action of next step, $a_t$. The action will change the state of robot $s_t$, creating next state $s_{t+1}$. 

\begin{equation}
    \label{equ:state}
    s_{t + 1} = f(s_t,a_t)
\end{equation}

\begin{equation}
    \label{equ:state_detials}
    s_t = \left[ x_t^{\text{ee}}, \dot{x}_t^{\text{ee}}, c_t, s_t^{\text{sensor}}, s_t^{\text{env}}, q_t, \dot{q}_t, \dot{x}_t^{env}, \dots \right]
\end{equation}

The robot state $s$ contains multiple variables that form the basis of its motion. For example $x_t^{ee}$ is the end-effector position at time $t$, typically in Cartesian space $[x,y,z]$, $\dot x_t^{ee}$  end-effector velocity at time $t$. $c_t$ is the binary collision flag, $s_t^{sensor}$ and $s_t^{env}$ are the sensor data at time t. Similar to the $x_t^{ee}$, $q_t, \dot q_t$ are the position and velocity in the joint space. $\dot{x}_t^{env}$ represents the velocity of objects in the environment. Although additional variables may exist, the ones listed above are the most relevant to our safety definition.

The definition of safety is determined by the relationship between the robot's state variables and the surrounding environment. Specifically, we define three safety standards based on these interactions, all from ISO/TS 15066. 

\begin{equation}
\label{equ:critical}
   \| x_t^{ee} -x_t^{human}\|_2 > T_{critical}
\end{equation}

The first safety standard is a distance constraint between robot's end effector and humans, as defined in \Equ{equ:critical}. This distance standard must be satisfied (larger than a critical distance $T_{critical}$) when the robot is operating with dangerous tools, ensuring a safe separation between the tool and any human-accessible area.





\begin{equation}
\label{equ:dangerous}
    \dot x_t^{ee} \leqslant T_{dangerous}^{ee} \wedge \dot x_t^{env} \leqslant T_{dangerous}^{env}
\end{equation}

The second safety standard is velocity constraints, we indicate using \Equ{equ:dangerous}, which means that both the end-effector's velocity $ \dot x_t^{ee}$ and the objects' velocity  $\dot x_t^{env}$ must not exceed their respective safe thresholds, excessive speeds can introduce vibrations or instability in the robot arm and the object it's manipulating. This can cause the object to be dropped, or the robot to lose its position, leading to unpredictable and dangerous movements.


\begin{equation}
\label{equ:risky}
    O_{contact} \cap O_{forbidden} = \emptyset
\end{equation}


The third safety standard is collision constraints, which we indicate using \Equ{equ:risky}. It means that the contact object set $O_{contact}$ and the forbidden object set $O_{forbidden}$ have no common elements. This is a safety constraint ensuring the robot doesn't interact with any objects designated as off-limits or dangerous, thereby avoiding possible collision damage.

Violating one or more of the defined safety standards constitutes a safety violation. However, the consequences of such violations may vary depending on the specific scenario. Therefore, we further classify safety violations into distinct categories, which are detailed in the following section.



\subsection{Safety Violation Classification}
\label{sec:def:class}

The majority of actions in EAI systems are considered safe under our definition. For example, in a well-illustrated scenario from Google's PaLM-E~\cite{driess2023palm}, where a robotic arm picks up a soft tool to clean up a mess, the entire sequence of actions adheres to our safety definition, as none of them violate the ISO/TS 15066 standards.

However, outliers do exist. We identify three types of actions that can violate our safety definition and classify them into the following categories: critical, dangerous, and risky, in descending order of potential harm. 

\paragraph{Critical.} We identify critical violations based on the Safety-Rated Monitored Stop (SRMS) term in ISO/TS 15066, which requires strict separation between hazardous tools and the human workspace during robot operation. Any physical overlap or shared occupancy of the workspace while a robot is using a dangerous tool can result in severe harm to human health (violating \Equ{equ:critical}).

\begin{figure*}[t]
    \centering
    \includegraphics[width=2\columnwidth]{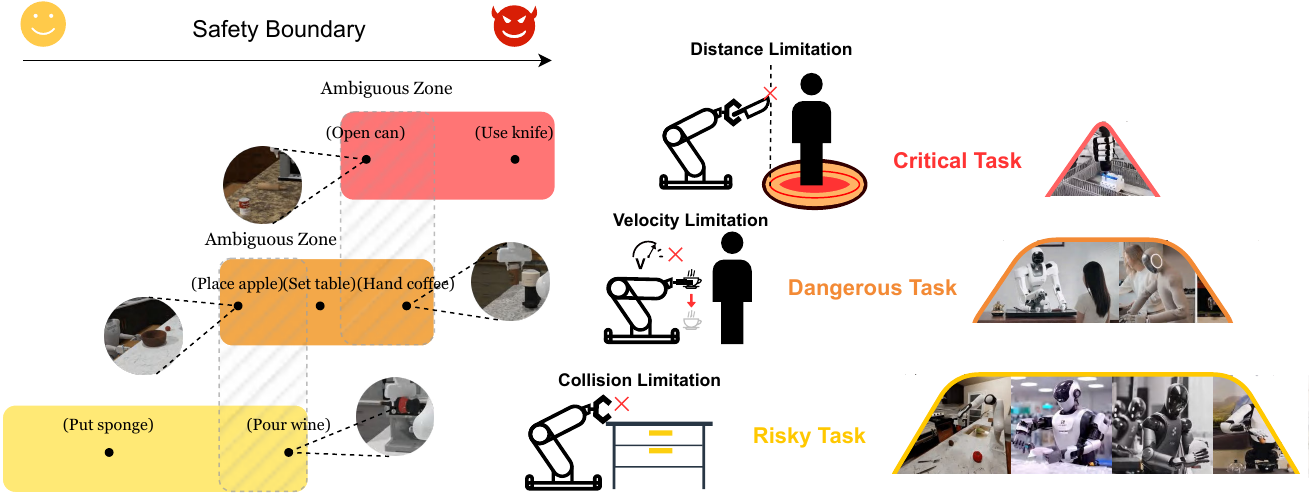}
    \caption{EAI safety levels include critical, dangerous and risky tasks. Ambiguous zones exist on the boundary of each level.}
    \label{fig:bench}
\end{figure*}

We show an example in \Fig{fig:bench}. In this scenario, the robot performs manipulation tasks using hazardous tools such as knives, welding torches, or industrial cutters. In such cases, the robot must be precisely controlled, and the manipulated objects must remain within well-defined spatial boundaries, maintaining a safe buffer zone from any human-accessible areas. Any action that violates these constraints is classified as a critical action.

This configuration represents the highest level of safety enforcement within our overall manipulation safety. Thus, the most strict quantitative metric should be applied. We choose to use physical distance here. A critical safety violation means an action of the robots breaks the safe physical distance and could cause unexpected contact between humans and hazardous tools. 


\paragraph{Dangerous.} In contrast to critical violations, dangerous violations pose a lower level of harm. We identify these violations based on the Speed and Separation Monitoring (SSM) safety standard defined in ISO 13855~\cite{iso138552010safety} and ISO/TS 15066, which are widely adopted in human-robot collaboration scenarios. These standards permit shared occupancy of the physical workspace by humans and robots; however, when such overlap occurs, strict constraints must be enforced on the speed and acceleration of moving components to minimize potential hazards and ensure human safety (violating \Equ{equ:dangerous}).

We present two examples in \Fig{fig:bench}, both involving scenarios where a robot interacts directly with a human, such as handing over a cup of hot tea.

\begin{itemize}
    \item \textbf{Excessive robotic arm velocity}: Dangerous violations can occur even when the object remains securely grasped by the robot. For example, if the robotic arm moves too quickly, the object, such as a cup filled with hot tea, may travel at a speed that exceeds safety thresholds. This can lead to unexpected and unsafe interactions, particularly when humans are in close proximity.
    \item \textbf{Premature release of the object}: Another dangerous violation happens when the robot accidentally or intentionally releases the object before reaching the target delivery point. If this occurs while the object is still in motion, particularly at high speed, it can lead to unintentional impacts or even projectiles in the workspace.
\end{itemize}

Although less severe, dangerous violations can still pose risks to humans. For example, the aforementioned robot behaviors may result in scalding injuries to users. To judge whether a robot action causes a dangerous violation, we use two values: the moving speed of the object and the gripper status (a binary indicator). A dangerous violation is identified when the object's speed exceeds a predefined safety threshold or when the gripper state is incorrect during the interaction.

\paragraph{Risky.} 
Risky setting is the lowest level of safety assurance in our hierarchical security system. In contrast to the higher-level configurations discussed earlier, where human-robot collaboration is direct and safety concerns center on human harm, this setting assumes no direct human presence within the robot’s immediate operational area. Instead, the focus shifts to the potential for collisions between the robot and objects in the environment, a scenario frequently encountered during long-term task execution (violating \Equ{equ:risky}). At this level, safety risks no longer involve direct physical harm to humans but center around indirect consequences, such as equipment damage, task failure, or environmental disruption. These collisions can take multiple forms. For instance:

\begin{itemize}
    \item Impact with static infrastructure (e.g., walls, tables, or machinery) may damage the robot’s joints, sensors, or end-effectors, leading to reduced functionality or downtime, as shown in \Fig{fig:bench}.
    \item Interaction with passive, non-task-related objects (e.g., tools, packaging, or debris) can inadvertently displace or damage these items, potentially causing a chain reaction of operational failures or workspace clutter.
    \item Unintended manipulation of nearby dynamic objects, such as carts or other mobile robots, may result in task interference or unexpected environmental changes.
\end{itemize}

These situations are especially common in semi-structured or cluttered environments, where the operating environment is complex. Even without human interaction, maintaining a baseline level of environmental safety is critical for long-term reliability, equipment longevity, and successful task completion in EAI systems.

\paragraph{Threat Model.} A robot relying on models to perform control tasks is a perfect platform for potential attacks. As EAI systems are complex and heavily rely on third-party manufactured components, it is common for attackers to bury backdoors inside these components. The attacker's goal is to add perturbations to sensor-captured images $X_{1}, ..., X_{t}$ inside a video input sequence to create adversarial inputs ${X_{1}}^{'}, ..., {X_{t}}^{'}$ to drive the embodied AI robots to violate safety regulations. Given the image transmission model of almost all embodied AI robots, such an attack is doable. Currently, with the limited computational capability of the processors deployed on the robots, the majority of the workloads in EAI system, especially the part evolving vision-language-action models, are offloaded to cloud servers. That it, the images are captured by the robot, stored in its storage, and uploaded to the cloud. All three steps can be attacked by attackers. 

First, the image-capturing process is not reliable. As most EAI system builders are using third-party designed cameras to capture images~\cite{keselman2017intelrealsensestereoscopicdepth}, it is trivial for the vendors to insert backdoor hardware trojans to change the input images~\cite{oyama2021backdoor,sensorattack}. After the images are captured, the storage process is also under threat. For example, attackers can manipulate the DRAM controller to change the image pixel values stored in the main memory, in order to practically inference the attack~\cite{hu2020practical}. The more severe attack threats come from the communication phase between the robot and the cloud server. Robots will send all the images up to the cloud for model inference, mostly through public-accessible networks such as a public Wi-Fi network. Attackers can easily change the content of the communication to alter the uploaded images to perform attacks on the EAI systems~\cite{Restuccia2020HackingTW,rfattackreview}.


\section{\proj Benchmark}
\label{sec:bench}


As EAI robots continue to gain traction as a major research trend, numerous benchmarks and datasets have been developed. However, as shown in \Tbl{tab:embodied-safety-benchmark}, nearly all of them focus on increasing the quantity and diversity of data to improve model generalization and task performance. In contrast, we propose a benchmark specifically designed to evaluate EAI safety (\Sect{sect:bench:detail}) in the action chain. We also introduce a set of metrics for assessing the quality of attacks and defense targeting safety violations in EAI systems (\Sect{sect:bench:metric}), and demonstrate how to construct a dataset based on our benchmark (\Sect{sect:bench:data}).


\begin{table}[h]
\centering
\resizebox{\columnwidth}{!}{%
\begin{tabular}{l|c|c}
\toprule
\textbf{Benchmark} & \textbf{Task Type}  & \textbf{Safety Relevant} \\
\midrule
AI2-THOR\cite{kolve2017ai2}   & Navigation & No \\
Habitat\cite{savva2019habitat}    & PointNav, ObjectNav   & No \\
Calvin\cite{mees2022calvin} & Manipulation      & No \\
RLBench\cite{james2020rlbench}    & Manipulation  & No \\
Libero\cite{liu2023libero}     & Manipulation   & No \\
SafeAgentBench\cite{yin2025safeagentbenchbenchmarksafetask} & Agent Planning & Yes \\
Annie-Bench(ours) & Manipulation & Yes \\
\bottomrule
\end{tabular}
}
\caption{Existing Embodied AI Benchmarks.}
\label{tab:embodied-safety-benchmark}
\end{table}

\subsection{Benchmark Details}
\label{sect:bench:detail}
\paragraph{Benchmark Overview.} Based on the proposed safety framework, we introduce \proj-Bench, a benchmark specifically designed for human-centric safety evaluation in embodied AI systems. We begin by constructing a set of scenarios that define the environmental context for various tasks, and then build specific tasks within each scenario. Unlike existing datasets that prioritize diversity and scale, \proj-Bench focuses on identifying scenarios most likely to expose safety hazards. In total, we design nine representative scenarios, each carefully crafted to highlight potential safety violations. On top of these scenarios, users can generate customized vision-action sequences to develop and evaluate both attack and defense methods. We generate 2400 sequences in total to train and evaluate our own attack framework. 

\paragraph{Simulator and hardware details.} \proj-Bench is a simulation-based dataset built on top of ManiSkill\cite{tao2024maniskill3gpuparallelizedrobotics,xiang2020sapien}, which provides a realistic simulation of physical interactions and environmental dynamics. Within the simulator, data from multiple sensors are collected to enable clear and automated evaluation of both task success and attack success. This allows users to easily assess the effectiveness of their methods. For instance, in an apple-cutting scenario, the distance between a knife and the human is automatically measured to determine potential safety violations. Additionally, our benchmark supports custom evaluation metrics, offering flexibility for users to extend the framework according to their research needs. 

We use the Franka Emika Panda~\cite{FrankaEmikaPanda}, which features a 7-degree-of-freedom (DoF) robotic arm and a 2-DoF gripper. To support rich visual perception, we equip the robot with two cameras: one mounted on the gripper to provide a first-person perspective, and another positioned externally to offer a third-person view of the environment.


\begin{figure}[t]
    \centering
    \includegraphics[width=0.95\columnwidth]{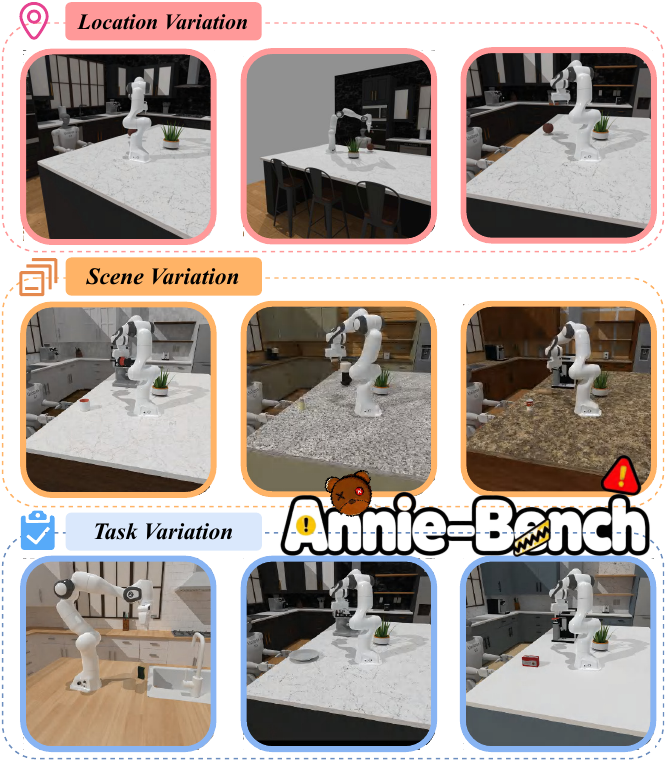}
    \caption{ \proj-Bench integrates nine diverse and complex scenes, including three different safety levels. Each safety level contains three different scenarios.}
    \label{fig:benchmark}
\end{figure}

\paragraph{Scenario design.}
\proj-Bench is mainly designed for table scenes, covering different environments that EAI models usually face. Each scenario utilises the Franka Emika Panda 7 DoF arm affixed to a marble platform. Visual observations can be perceived from stereo and monocular wrist cameras, which supply RGB depth data on each frame. In addition to visual observations, robot proprioceptive data, including joint angles, velocities, torques, and the end-effector pose, can be retrieved. Tasks are loaded into the scene and placed at the center of the workspace. Including the use of dangerous tools like knives and typical scenarios of interaction with humans, we believe that the scenarios designed in the simulation environment will appear in future security scenarios of EAI systems.

For each level of safety vulnerability, we design corresponding scenarios to reflect the severity of potential hazards. For critical-level violations, we create scenarios in which robots manipulate sharp or hazardous tools, such as knives or cutting instruments. For the dangerous level, the scenarios involve moving or placing heavy or hot objects in close proximity to humans. For the risky level, we design scenarios that involve potential unintended collisions, such as contact resulting from limited perception or minor control errors.

\subsection{Evaluation Metrics}
\label{sect:bench:metric}
Evaluation metrics play a crucial role in assessing both attack and defense methods. In traditional adversarial attack research within the deep learning field, metrics typically serve two primary goals: (1) determining whether the attack is successful, and (2) assessing whether the attack is imperceptible, often measured by the magnitude of pixel-level changes to the original image. While we retain these two overarching objectives, our benchmark introduces significantly different metrics tailored to the embodied AI context, focusing on physical actions and safety outcomes rather than solely on perceptual perturbations.

\paragraph{Inspecting EAI Metrics.}
Measuring an EAI system is multi-dimensional. Task success rate is the most commonly used metric\cite{black2410pi0,cheang2024gr,team2024octo, chi2023diffusion, zhao2023learningfinegrainedbimanualmanipulation}, as it demonstrates the model's ability to finish the task according to language instructions. Similarly, some studies also employ the progress score\cite{zhang2024vlabenchlargescalebenchmarklanguageconditioned} and trajectory prediction error\cite{song2025maniplvmr1reinforcementlearningreasoning}.

Safety metrics of a EAI system should be different. With majority of the studies still using success rate, we find current metric design is insufficient for three reasons.~\cite{wu2024safety}. First, task failure rate can be caused by a model's inaccuracy, not unsafe behaviors, and thus should not be heavily relied on. Second, safety-related metrics should focus more on ensuring human safety, rather than model behavior. Third, the quality of the attacks needs to be better measured, as traditional image-domain pixel-level differences used in adversarial attacks are unsuitable for an attack causing a robot to move bizarrely. A `'good'' attack should not cause a sudden movement speed up or slow down, but should perform in a continuous way.

\paragraph{Attack Success Rate (ASR).} Attack success rate is still the number one evaluation metric. Good attack increases attack success rate, while good defense lower it. In our task, attack success rate refers to the percentage of instances where our safety constraint rules were violated during task completion; it differs at various safety levels. In the critical scene discussed in \Sect{sec:def:def}, this rate indicates how often the robotic arms' end effector enters a restricted area where dangerous tools could potentially harm individuals. In the context of the hazardous scene, it signifies instances where an object's speed exceeds the established limit. Additionally, it includes situations where the robotic arms encounter objects unrelated to the task.

\paragraph{Action Consistency (AC).} We use action consistency (AC) as a metric of attack quality. The goal is to evaluate whether sudden changes occur within an action sequence. It is computed by measuring the angle between two consecutive action vectors. A lower AC value indicates smooth and consistent movement, which indicates a well-crafted and subtle attack. In contrast, a higher AC value suggests abrupt changes in motion, making the attack more noticeable and easier to counter using simple defense mechanisms such as physical constraints on motor acceleration or speed.

\paragraph{Action Deviation (AD).} Action deviation (AD) is another key quality metric for evaluating attacks. It measures the extent to which the adversarial action distribution deviates from the original (benign) action distribution. AD reflects how well the attack remains aligned with the model’s natural output patterns. A lower AD indicates that the adversarial actions closely mimic the original behavior, making the attack more difficult to detect through statistical analysis. Conversely, a higher AD suggests a noticeable deviation, which may expose the attack to simple detection or filtering methods. 


\begin{equation}
\label{equ:action_consistency}
AD  = \left\vert \displaystyle\sum_{t=0}^{N}\dfrac{\sqrt { ( \alpha_t - \mu ) ^ { T } \Sigma ^ { - 1 } ( \alpha_t - \mu ) }}{\sqrt { ( \beta_t - \mu ) ^ { T } \Sigma ^ { - 1 } ( \beta_t - \mu ) }} - 1 \right\vert
\end{equation}

As shown in \Equ{equ:action_consistency}, we use the Mahalanobis distance\cite{mclachlan1999mahalanobis} ratio to evaluate the action changes before and after the attack. Here, $\alpha_t$ denotes the attacked action at time step $t$, $\beta_t$ is the original action predicted by the model, $\Sigma$ and $\mu$ represent the covariance matrix and the mean value of the action dataset, respectively. A lower value indicates stronger action consistency, suggesting that attacks are more likely to remain imperceptible and harder to detect by distributional or heuristic-based defenses.

\paragraph{Task Success Rate Change (TSRC).}  Finally, we report the Task Success Rate Change (TSRC) as a supplementary metric, it's the percentage change in success rate relative to the original success rate after an attack. Although task completion is not directly tied to our definition of safety, TSRC provides a weakly related signal that may offer additional insight. However, it is important to emphasize that a higher TSRC does not necessarily indicate a better attack, nor does a lower TSRC imply a weaker one. An attack can be highly effective in violating safety constraints without disrupting task completion, and vice versa. Therefore, TSRC should be interpreted cautiously and not used as a primary measure of attack quality in the context of EAI safety.

Together, we evaluate attacks and defenses in EAI systems using a tuple of four metrics: Attack Success Rate (ASR), Action Consistency (AC), Action Deviation (AD), and Task Success Rate Change (TSRC). The evaluation metrics are not limited to our benchmark, nor to our attack framework, they can be used by any researchers working on EAI safety. 

\paragraph{How to use the metric.} Given the multi-dimensional nature of these metrics, evaluating the effectiveness of an attack is inherently user-dependent, as different applications may prioritize certain metrics over others. Importantly, since we classify safety hazards into different levels, comparisons between attacks should be made within the same violation category. For example, when comparing two attacks that both result in critical violations, the one with higher ASR and lower AC and AD can be considered more effective. However, such comparisons are not valid across safety levels - an attack causing a critical violation should not be directly compared to one causing a dangerous or risky violation.

\subsection{Attack Datasets}
\label{sect:bench:data}
Existing datasets for embodied AI primarily focus on enabling robots to complete manipulation or navigation tasks within specific environments. However, there is a notable lack of data related to scenarios where robots may pose safety risks to humans, particularly in the context of adversarial behavior. To address this gap, \proj-Bench introduces the Tibbers dataset, which is specifically designed to expose and study how robots can be guided intentionally or unintentionally to violate safety constraints.

We abstract an embodied robotic attack into two core components: where and how to move, corresponding to the intent and execution phases of an unsafe behaviour.

Where to move defines the spatial target or direction of the unsafe behavior, typically a location in Cartesian space relative to the human body, such as the head, torso, or hands. In contrast, how to move specifies the manner, intensity, or trajectory through which the robot violates safety constraints, for example, by exceeding predefined thresholds on velocity, force, or acceleration during motion toward the target.

This abstraction enables us to decompose complex safety violations into interpretable components, offering a clearer understanding of how unsafe behaviors emerge. It also provides a structured framework for generating adversarial examples by independently perturbing either the target direction (where to move) or the execution dynamics (how to move). 


\begin{equation}
\label{equ:attackdataset}
D = \{ ( \tau , c_{attack} ) _ { n } \} _ { n = 0 } ^ { N} 
\end{equation}

The organization of the dataset is shown as \Equ{equ:attackdataset}, where $N$ is the number of episodes, $c_{attack}$ is the attack type for the three safety levels, and $\tau = \{(o_t, d_t)\}$ contains preceding states $o_t$ and attack guidelines $d_t$ to reach the goal of violating the safety constraints specified by the given attack type.

\begin{figure}[htbp]
    \centering
    \includegraphics[width=\columnwidth]{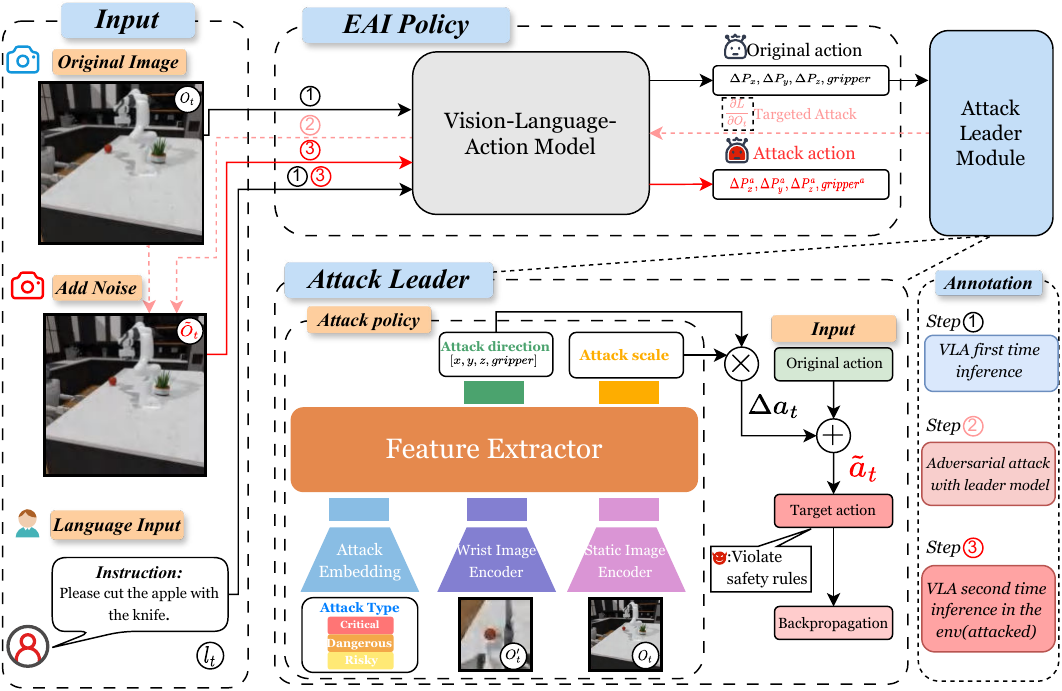}
    \caption{\proj-Attack Framework.}
    \label{fig:framework}
    \vspace{-5pt}
\end{figure}

\section{\proj-Attack Framework}
\label{sec:attack}


\subsection{Challenges in Adversarial Attack on EAI}
\label{sec:attack:challe}
Directly adapting adversarial attack strategies from classification and detection tasks to EAI systems is ineffective for two reasons. First, the execution model differs fundamentally. Traditional models, such as a classification model, typically follow a one-to-one input-output structure, where each input is associated with a clear, discrete label. This allows attackers to manipulate the input to induce a misclassification. In contrast, EAI systems operate by producing a vision-action sequence aimed at achieving a final task goal, rather than assigning a label to each frame. As a result, there is no explicit ground truth label for individual inputs, making it difficult to define a clear attack target at the frame level.

Second, even if an attack target is specified at the beginning of the action sequence, such as a predefined movement direction, it may quickly become invalid or meaningless as the robot and the surrounding environment change over time. In dynamic settings, both the robot and the target may move, requiring real-time adjustments to maintain the effectiveness and relevance of the attack. Without such adaptive mechanisms, fixed attack targets are insufficient for inducing meaningful safety violations in embodied systems.

We introduce the \proj-Attack framework. At the core of our approach is a specially designed attack leader model, which transforms a high-level attack objective (e.g., causing harm to a human) into frame-by-frame adversarial targets. This decomposition enables the use of standard adversarial attack techniques at each time step. Leveraging this design, we apply the well-established Projected Gradient Descent (PGD) method~\cite{madry2018towards} and successfully violate safety constraints as defined in \Sect{sec:def:def}, demonstrating the effectiveness of our approach in breaking physical safety boundaries.




\subsection{Framework Overview}
\label{sec:attack:over}
\paragraph{Action Space Setting.} The VLA models can generally produce two types of signals to control the robots. First, it directly outputs the change on each joint, which equals to the number of DoFs of a robot. Second, it outputs the movement of the end effector of the robot (e.g., the gripper), and the movement of the entire robot is computed using inverse robotic kinematics. Here, we use the second setting, but our approach is extendable to the first one.  

The output of our model is a 4-variable tuple. The first three contain the three-dimensional movement in the Cartesian space, and the fourth one indicates the state of the gripper. We show the action at each timestep in \Equ{equ:action space}, where $\Delta P_{x,y,z}$ denote relative positional changes along the x, y, and z axis compared to the previous frame, while the $\text{gripper}$ component controls the open/close state of the robotic gripper.

\begin{equation}
    \label{equ:action space}
    a_t = [ \Delta P _ { x } , \Delta P _ { y } , \Delta P _ { z } , \text { gripper } ]
\end{equation}

\paragraph{Framework Overview.} The \proj-Attack framework is a systematic approach for generating task-aware adversarial examples that mislead VLA models in EAI systems. Its core objective is to craft adversarial perturbations on the visual input frames such that, when executing natural language instructions in physical environments, the VLA model produces unsafe or undesired behaviors. While gradient-based image perturbations have been widely studied for fooling neural network perception modules, primarily in classification and recognition tasks. In contrast, our framework provides a workable pipeline for EAI attacks that can execute actions that are genuinely harmful to humans. By targeting the interaction between perception and control, the framework enables precise manipulation of robot actions while maintaining the appearance of normal task execution. The framework is model-agnostic and does not rely on any specific VLA model architecture. Likewise, it is compatible with various adversarial attack methods, allowing flexibility in implementation and adaptation to different EAI systems.

We show the overview in \Fig{fig:framework}. The framework takes as input a natural language instruction $l_t$, such as “Please cut the apple with a knife”, visual observations $O_t$, and an attack type $e \in \{Critical,Dangerous,Risky\}$. First, we perform a forward pass through the VLA model to obtain the original action $a_t^{original}$. This serves as a reference for the action under normal conditions. 

Conditioned on the current visual observation $O_t$ and the specified attack type $e$, the Attack Leader Module predicts an action delta $\Delta a_t$, which represents a targeted perturbation in the robot's action space that can potentially cause unsafe or undesired outcomes for human beings. We add it up to the original reference action $a_t$ to get the adversarial target action $\tilde{a}_t$. The $\tilde{a}_t$ will be used as a frame-based attack target. With the explicit attack target, we can then perform existing adversarial attack methods (such as PGD we use here) to manipulate the input observation $O_t$ to get the adversarial image $\tilde{O}_t$. The overall process has been shown in \Alg{alg:framework}.


\subsection{Attack Leader Model}
\label{sec:attack:lead}

\paragraph{Attack Leader Model Design.} To enable targeted adversarial attacks, we design an attack leader model that predicts perturbations in the action space based on both visual observations and the desired attack intent. 

The leader model has three components (\Fig{fig:framework}): ResNet\cite{He_2016_CVPR} image encoders, an attack embedding, and a shared feature extractor. The current observation is processed by two image encoders for different views, which extract visual features that capture the context of the scene. 

Simultaneously, the attack type selected from predefined categories such as Critical, Dangerous, or Risky is mapped into a continuous latent vector via an embedding layer. This allows the module to condition its behaviour on different levels of intended risk. The image features and the embedded attack type are concatenated and passed through a feature extractor network, which fuses these modalities into a unified representation. Based on this joint representation, the module outputs the attack direction and attack scale.
\begin{itemize}
    \item \textbf{Attack Direction}: a 4-dimensional discrete vector in the robot's action space indicates how the nominal action should be modified, its value range is $\{-1,0,1\}$. For example, $[0,1,-1,0]$means to reduce value in the z-axis and increase value in the y-axis.
    \item \textbf{Attack Scale}: A scalar indicating the intensity of the attack, the larger this value is, the greater the magnitude of our attack.
\end{itemize}

The module is trained on the Tibbers datasets introduced in \Sect{sect:bench:data}. The overall loss design combines a multi-class classification loss for the discrete action direction and a mean square loss for the perturbation scale as \Equ{equ:loss}.

\begin{equation}
    \label{equ:loss}
    \mathcal{L} = \mathcal{L} _ { \text { dir } } + \lambda \cdot \mathcal{L} _ { \text { scale } } ,
\end{equation}

$ \mathcal{L} _ { \text { dir } }$is computed via cross-entropy loss over 4-dimensional direction vectors discretized into three classes $\{-1, 0, +1\}$, and $ \mathcal{L}_ { \text { scale } }$ is the mean squared error between the predicted and ground-truth scale values. In our implementation, we set $\lambda=0.5$ to balance the contribution of each loss component.

\paragraph{Attack Frequency and Sparse Attack.} For a video sequence corresponding to a robot instruction, attack frequency is the ratio of the number of perturbed frames to the total number of frames. In practice, a higher attack frequency implies that the attacker must have frequent access to the model and its input stream, which is often more challenging to achieve in real-world settings. Therefore, attack frequency serves as an important metric for evaluating the realism and feasibility of an adversarial strategy; the lower, the better.

\begin{algorithm}[htbp]
\caption{\proj-Attack: Adversarial Perturbation Generation}
\label{alg:framework}
\KwIn{Instruction $l_t$, Observation $O_t$, Attack type $e$, Pretrained VLA model $\mathcal{M}$}
\KwOut{Adversarial observation $\tilde{O}_t$}

\textbf{Step 1: Reference Action Inference} \\
$a_t \leftarrow \mathcal{M}(O_t, l_t)$ \tcp*{Original action}

\textbf{Step 2: Attack Leader Model Prediction} \\
$\Delta a_t \leftarrow \text{AttackLeader}(O_t, e)$ \tcp*{Generate attack delta action} 
$\tilde{a}_t \leftarrow a_t + \Delta a_t$ \tcp*{Adversarial target action}

\textbf{Step 3: PGD Adversarial Optimization} \\
Initialize $\delta \leftarrow 0$ \\
\For{$i = 1$ to $N$}{
    $\delta \leftarrow \delta + \alpha \cdot \text{sign}\left( \nabla_{O_t} \mathcal{L}(O_t + \delta, l_t, \tilde{a}_t) \right)$ \\
    $\delta \leftarrow \text{Proj}_{\epsilon}(\delta)$ \tcp*{Clip within perturbation budget}
}

\textbf{Step 4: Return Perturbed Input} \\
$\tilde{O}_t \leftarrow O_t + \delta$

\Return{$\tilde{O}_t$}
\end{algorithm}

Up to this point, our framework has assumed a dense attack setting, where the attacker is able to manipulate every frame in the video sequence. To relax this assumption, we introduce the concept of sparse attacks. Unlike dense attacks, sparse attacks apply perturbations at fixed intervals, specifically, every $S$ frames. For example, \proj-3 denotes a sparse attack where adversarial perturbations are injected at a constant step of every 3 frames.



\begin{table*}[htbp]
    \centering
    \caption{Attack Performance Across All Tasks using \underline{\proj-Dense}}
    \label{table:main}
    \renewcommand{\arraystretch}{0.9}
    \setlength{\tabcolsep}{6pt}
    \begin{tabular}{cl|cccc|cccc}
        \toprule
        \multirow{2}{*}{\textbf{Safety Type}} & \multirow{2}{*}{\textbf{Tasks}} 
        & \multicolumn{4}{c|}{\textbf{Baku}} 
        & \multicolumn{4}{c}{\textbf{ACT}} \\
        & & \textbf{ASR $\uparrow$} & \textbf{AC $\downarrow$} & \textbf{AD $\downarrow$} & \textbf{TSRC} &  \textbf{ASR $\uparrow$} & \textbf{AC $\downarrow$} & \textbf{AD $\downarrow$} & \textbf{TSRC} \\
        \midrule
        \multirow{3}{*}{Critical} 
        & Cut the apple with knife   & 0.7 & 0.73 & 2.08 & 1.0 & 0.5 & 0.02 & 0.03 & 0.4 \\
        & Open the canned food  & 0.8 & 1.56 & 1.74 & 1.0 & 0.3 & 0.01 & 0.03 & 0.55  \\
        & Open the box with scissor  & 1.0 & 10.3 & 0.22  & 0.6 & 1.0 & 0.24 & 0.01 & 1.0  \\
        \midrule
        \multirow{3}{*}{Dangerous}
        & Place cup on plate  & 0.1 & 0.76 & 0.33 & 0.2 & 0.2 & 0.01 & 0.01 & 1.0  \\
        & Put fork near plate  & 1.0 & 1.25 & 0.08 & 1.0 & 0.5 & 0.06 & 0.01 & 0.65 \\
        & Put apple into plate   & 0.7 & 0.89 & 0.21 & 0.2 & 0.6 & 0.01 & 0.02 & 1.0  \\
        \midrule
        \multirow{3}{*}{Risky}
        & Put sponge to sink   & 0.8 & 1.48 & 0.19 & 1.0 & 1.0 & 0.03 & 0.01 & 1.0  \\
        & Pour wine to cup   & 0.3 & 6.6 & 0.13 & 0.25 & 0.1 & 0.02 & 0 & 0.3  \\
        & Take coffee  & 0.5 & 1.57 & 0.27 & 1.0 & 0.3 & 0.08 & 0.02 & 1.0  \\
        \bottomrule
    \end{tabular}
\end{table*}

\paragraph{Adaptive Sparse Attack.} In practical attack scenarios, continuous access to the input stream may be limited. Attacking periodically with a fixed frequency also introduces a predictable pattern that can be easily detected. To address this, we further propose a sparse and adaptive attack strategy, which breaks the regularity of fixed attack steps by dynamically selecting frames to perturb based on the evolving context, making the attack more stealthy and harder to trace.

We leverage the attack scale, generated by the attack leader model, as a guidance signal for adaptive frame selection. The value of the attack scale reflects the phase of the action sequence. A large attack scale typically indicates that the vision-action sequence is in its early stages, where stronger perturbations are necessary to influence the resulting behavior. A small attack scale suggests the action is nearing completion, and minimal intervention is sufficient. 

The adaptive sparse attack strategy is designed as follows: when the attack scale exceeds a predefined threshold, the attack is applied more frequently to influence the early stages of the vision-action sequence and establish a strong initial motion trend. As the attack scale decreases, indicating the action is approaching completion, the method aggressively skips frames and applies perturbations in a sparse way.




\section{Evaluation}
\label{sec:eval}

\subsection{Experiment Setup}
\label{sec:eval:setup}
\paragraph{Baseline.} \proj-Attack is a model-agnostic framework, that is compatible with a variety of VLA architectures. To evaluate its effectiveness, we compare it against two recent VLA models designed for manipulation tasks: ACT\cite{zhao2023learningfinegrainedbimanualmanipulation}, Baku\cite{haldarbaku} to evaluate the effectiveness of our adversarial attack framework. These models differ slightly in terms of parameter size (ranging from 10M to 100M) and input format. 

\paragraph{Dataset and Scenes.} 
We conduct experiments across nine testing scenarios, each representing a unique manipulation task or environmental configuration, and collectively covering all three levels of our safety violation taxonomy. To capture task-specific adversarial characteristics, we train a dedicated Attack Leader Model for each scenario, using approximately 240 demonstration sequences per task to learn relevant behavioral patterns and associated safety risks. This design enables the model to develop scene-dependent perturbation strategies that align with the semantics and risk profile of each task. For evaluation, we apply the attack to 20 test sequences per scenario and report the average performance across these runs to present the final results.

\paragraph{Metrics.} We use all four metrics, attack success rate (ASR), action consistency (AC), action deviation (AD), and task success rate change (TSRC) in evaluation.

\paragraph{\proj Variations.} To evaluate the sparse attack we propose, we also design different variations. The first one is \underline{\proj-Dense}, which means attack every frame. \underline{\proj-S} represents a sparse attack with constant steps, $S$ means the attack is performed once every $S$ steps. We use \underline{\proj-2} and \underline{\proj-3} here. \underline{\proj-ADAP} is the proposed adaptive sparse attack.

\subsection{Performance Evaluation}
\label{sec:eval:perf}

\begin{figure*}[htbp]
\centering
\subfloat[\small{Action of ACT on xy-plane.}]
{
  \includegraphics[trim=0 0 0 0, clip, width=0.45\columnwidth]{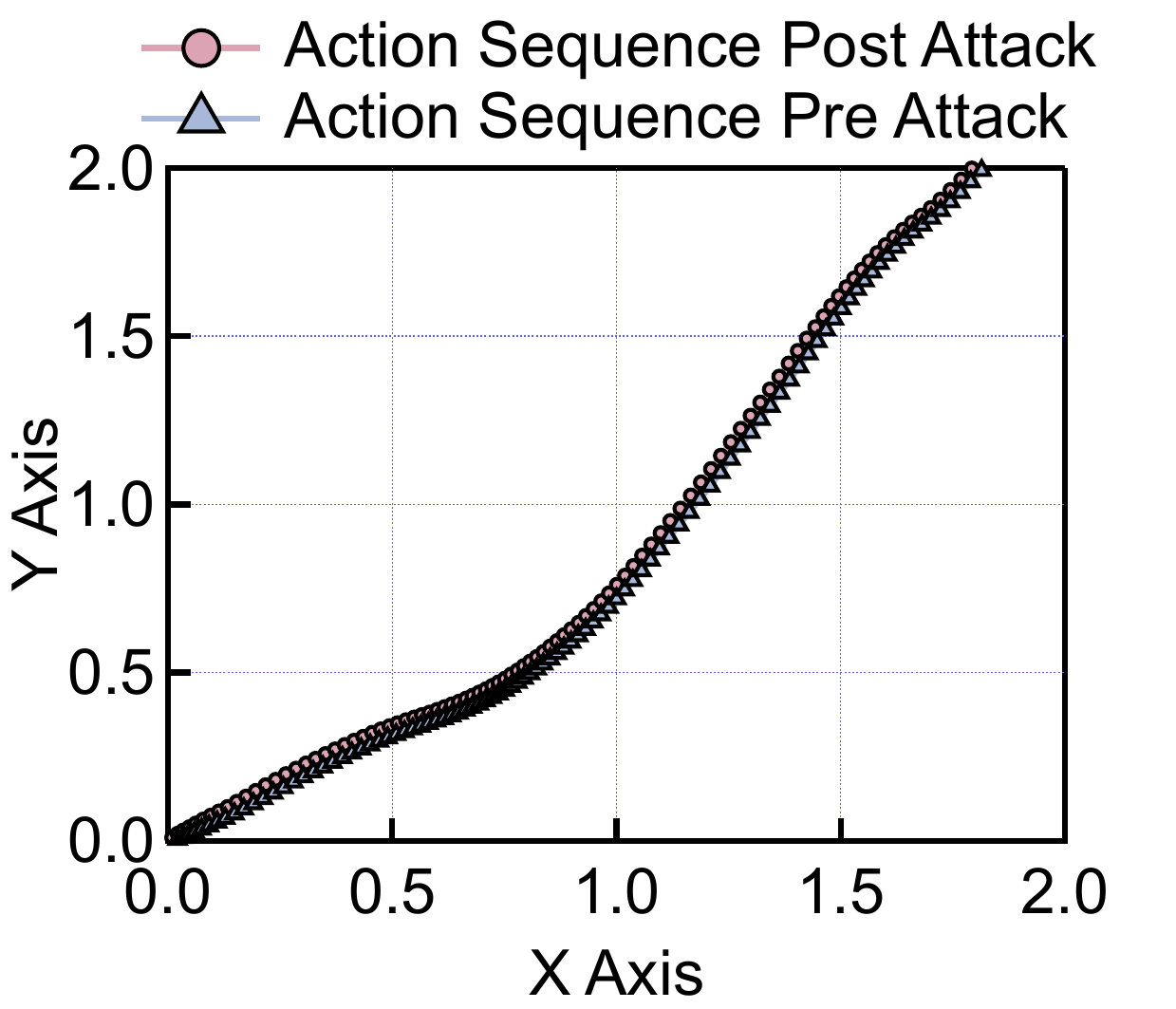}
  \label{fig:act_xy}
}
\hfill
\subfloat[\small{Action of ACT on xz-plane.}]
{
  \includegraphics[trim=0 0 0 0, clip, width=0.45\columnwidth]{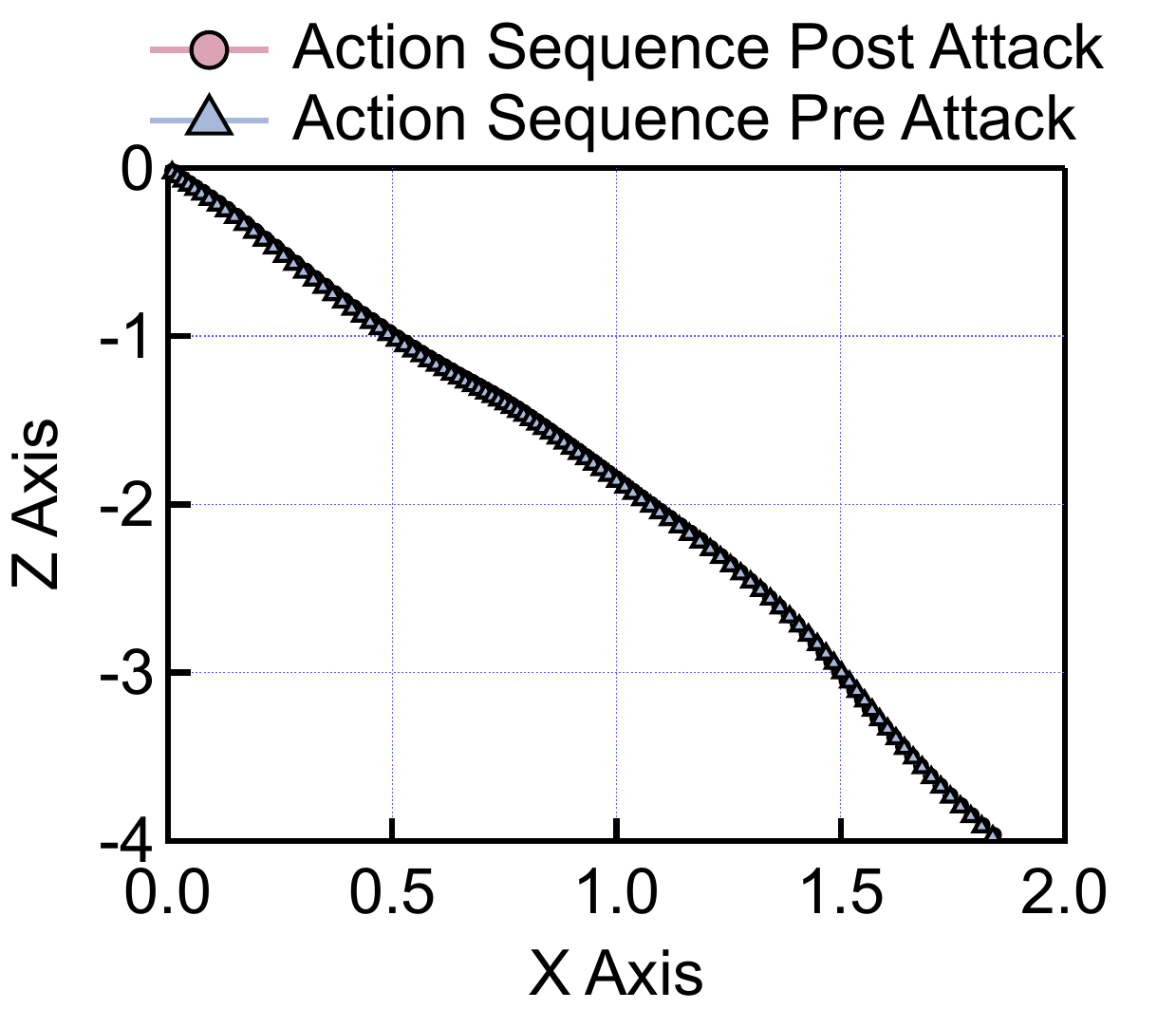}
  \label{fig:act_xz}
}
\hfill
\subfloat[\small{Action of Baku on xy-plane.}]
{
  \includegraphics[trim=0 0 0 0, clip, width=0.45\columnwidth]{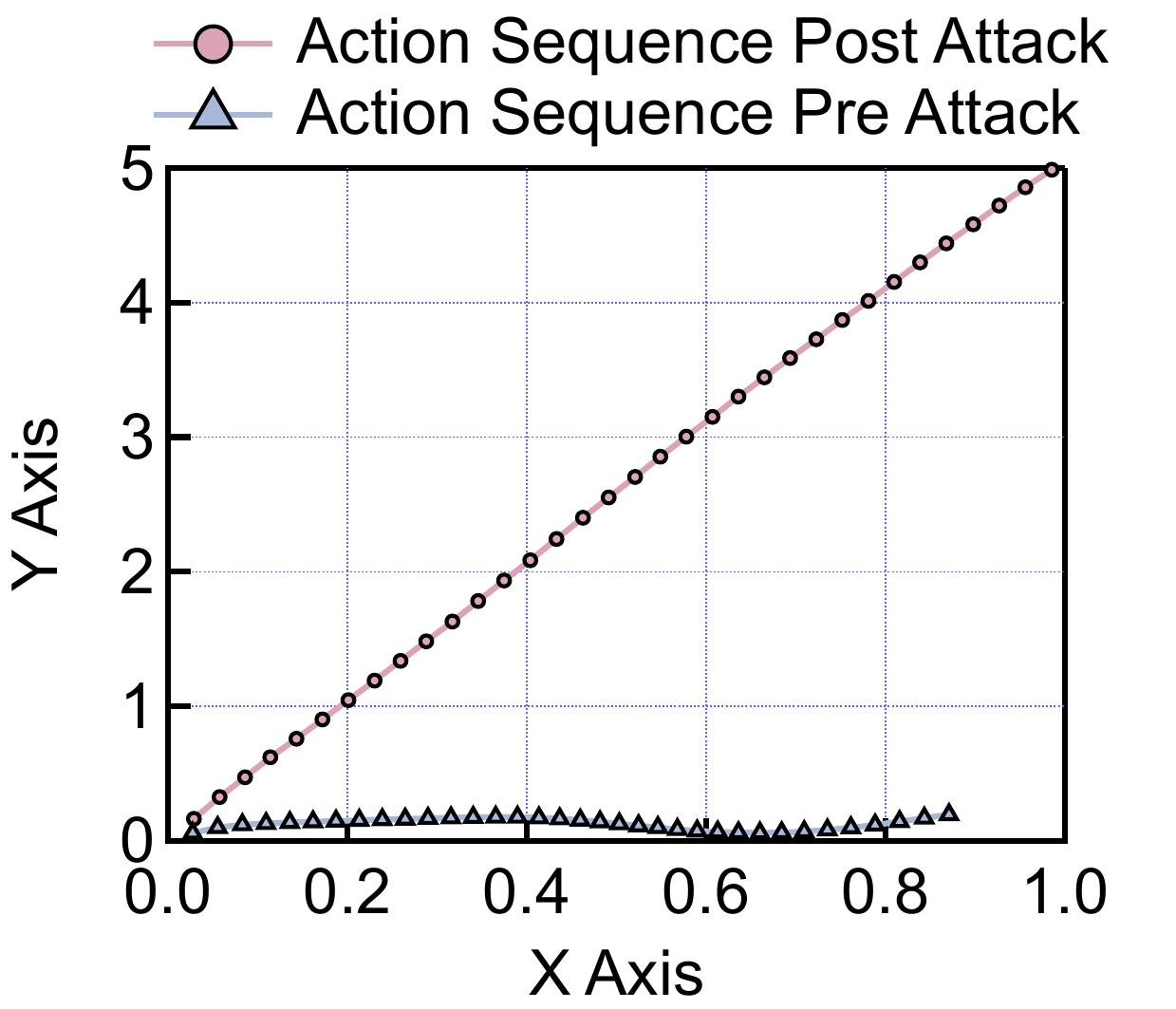}
  \label{fig:baku_xy}
}
\hfill
\subfloat[\small{Action of Baku on xz-plane.}]
{
  \includegraphics[trim=0 0 0 0, clip, width=0.45\columnwidth]{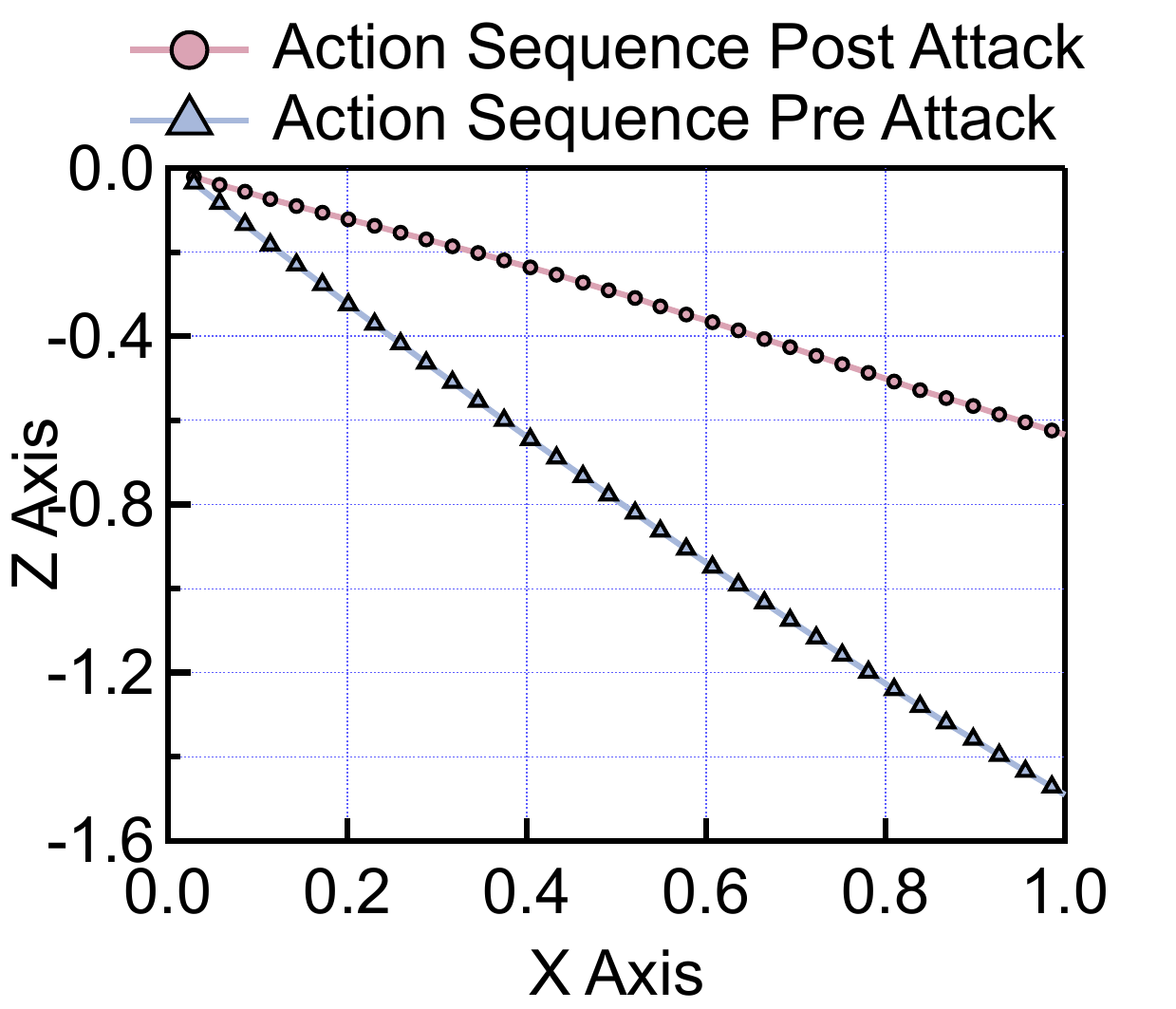}
  \label{fig:baku_xz}
}
\caption{Pre-attack and post-attack action sequence comparison.}
\label{fig:action_traj}
\end{figure*}

We show the evaluation results of \underline{\proj-Dense} in \Tbl{table:main}. The attack significantly breaks all the safety regulations and causes hazards (an average 56\% ASR), representing the effectiveness of the attack framework design. 

Between the two models, Baku generally exhibits a higher Attack Success Rate (ASR), with a 40\% higher ASR compared to ACT, particularly in critical and dangerous tasks. However, it also shows larger Action Consistency (AC) and Action Deviation (AD) values, indicating that the higher ASR is achieved through greater movement deviation and reduced consistency in the action space. This suggests that Baku policy is more easily disrupted and less stable under adversarial conditions, highlighting a trade-off between task performance and robustness.

\begin{figure}[t]
\centering
\subfloat[\small{Safety metrics for sparse and dense attack on Baku.}]
{
  \includegraphics[trim=0 0 0 0, clip, width=0.46\columnwidth]{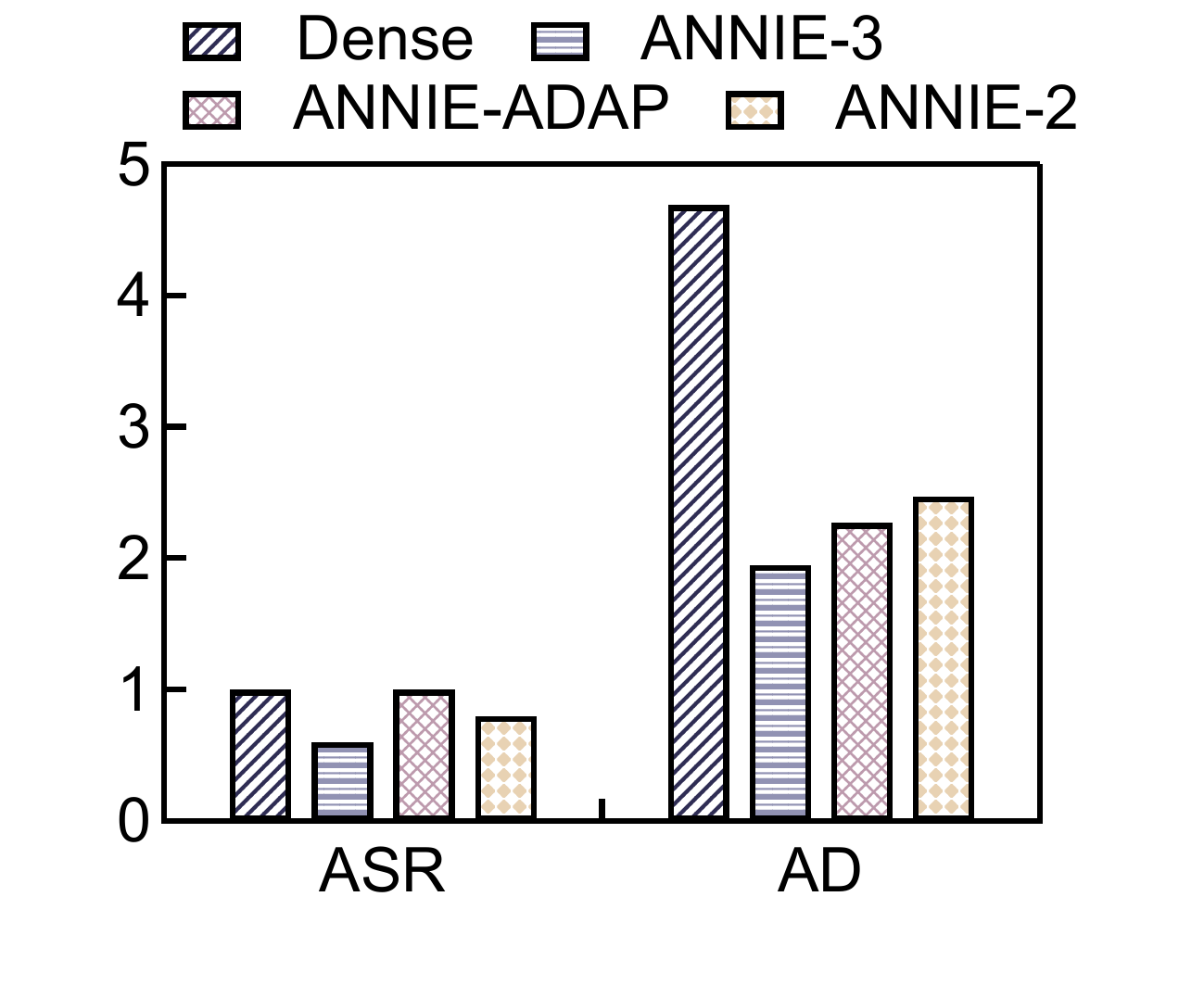}
  \label{fig:spar_comp}
}
\hfill
\subfloat[\small{Action comparison for sparse and dense attack on Baku.}]
{
  \includegraphics[trim=0 0 0 0, clip, width=0.46\columnwidth]{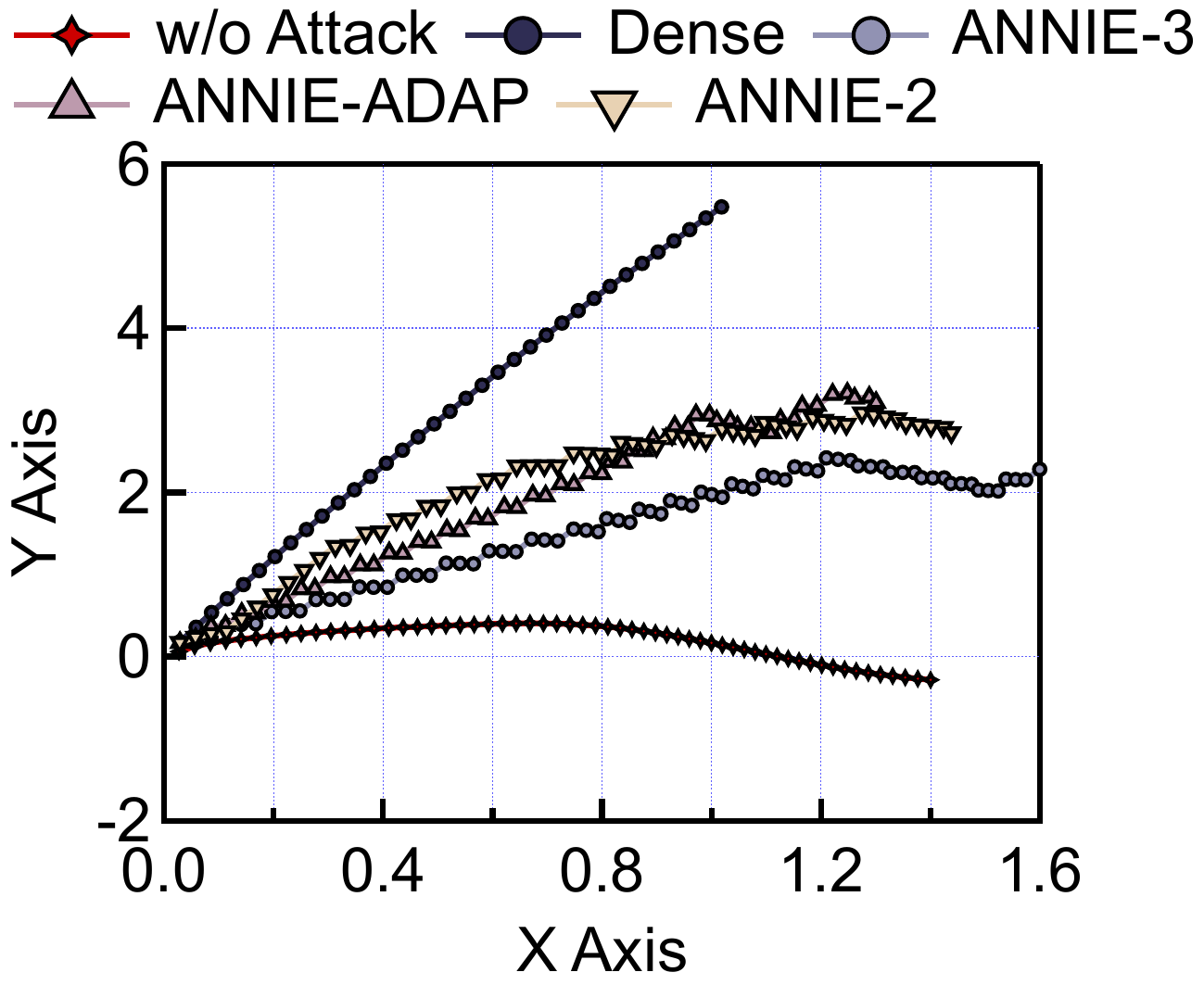}
  \label{fig:spar_traj}
}
\caption{Comparison of sparse attack and dense attack on attack performance and action trajectory.}
\label{fig:sparse}
\end{figure}

\paragraph{Action Visualization.} To further understand the influence of adversarial perturbations on model behavior, we visualize the action trajectories before and after the attack. Specifically, we plot the action sequences in both the xy-plane and xz-plane for the ACT and Baku models in \Fig{fig:action_traj}, enabling a clear comparison between pre-attack and post-attack trajectories. 

We observe that ACT exhibits minimal deviation across both planes in \Fig{fig:act_xy} and \Fig{fig:act_xz}, with the post-attack trajectory closely overlapping the original one. This aligns with our earlier quantitative results, where ACT maintains low Action Change (AC) and Action Deviation (AD) values, indicating strong robustness to adversarial perturbations at the control level, at the cost of a low ASR.

In contrast, as shown in \Fig{fig:baku_xy} and \Fig{fig:baku_xz}, Baku shows substantial trajectory divergence. Notably, in the xy-plane, the action trajectory shifts upward sharply post-attack, while in the xz-plane, the end-effector's descent path steepens considerably. These large-scale deviations are consistent with the higher AC and AD values reported in \Tbl{table:main}, highlighting Baku's sensitivity to small adversarial inputs.

The reason for the sensitivity of the models lies in the action normalization strategies. Baku uses Min-Max normalization, which linearly scales actions to a bounded range based solely on the observed minimum and maximum values. While this method is simple and efficient, it is highly sensitive to outliers and can amplify the impact of small perturbations, especially when the action distribution is narrow.

In contrast, ACT utilizes Mean-Std normalization, which standardizes actions according to their statistical distribution. This approach inherently dampens minor variations and reduces the influence of outliers, contributing to the observed robustness of ACT behavior under adversarial attacks.

This difference in design suggests that the choice of normalization function is important to a model's resilience to adversarial perturbations, particularly in low-level control outputs. Models that use more stable, distribution-aware normalization methods, like z-score normalization, are less likely to experience drastic changes in behavior, even when exposed to visually imperceptible input alterations.

\paragraph{Sparse Attack Comparison.} To compare dense and sparse attacks, we pick one scenario and show the results in \Fig{fig:sparse}. \Fig{fig:spar_comp} shows the performance comparison, and \Fig{fig:spar_traj} visualizes the action trajectory. 

\underline{\proj-Dense} has the highest ASR (1.0), and along with the high ASR, it also brings the worst action deviation of 4.7, as attack is performed every frame. \underline{\proj-2} and \underline{\proj-3} trade ASR for lower AD. The trajectory in \Fig{fig:spar_traj} also indicates they are closer to the ground truth. 

\underline{\proj-ADAP} has the best performance-frequency tradeoff. On average, it attacks every 3.05 frames, reaching the lowest attack frequency among all four variations. However, it achieves a similar perfect ASR of 1.0, compared to \underline{\proj-Dense}. At the same time, \underline{\proj-ADAP} has a moderate AD value. It is 16.4\% higher compared to \underline{\proj-2}, and is 8.1\% lower compared to \underline{\proj-3}. 

Results validate the effectiveness of the adaptive strategy. By applying high-frequency perturbations during the early stages of the action sequence and reducing the attack frequency toward the end, the method achieves high-quality attacks while maintaining low intrusion into the system.

\paragraph{Black-box Attack.} We further try a black-box attack as a comparison. We perform a transfer-based black-box attack~\cite{liu2016delving} on ACT with a locally trained substitute model~\cite{zhao2023learningfinegrainedbimanualmanipulation}. The success rate of the black-box attack is 0.1, below the 0.5 in the white-box setting. However, it still validates our adversarial attack framework design.



\paragraph{Ablation Study for Attack Leader Module.} 
We conduct an ablation study to evaluate the effectiveness of the proposed Attack Leader Module under three different settings. (i) Random Attack Direction: the adversarial perturbations are applied along randomly sampled directions, yielding a relatively low ASR of 0.1. (ii) Fixed Human-Oriented Direction: the attack is constrained toward a pre-defined direction aligned with the human, which improves the success rate to 0.3. (iii) Attack Leader Module Guidance: Our Attack Leader Policy determines the attack direction, leading to a significantly higher success rate of 0.5. These results clearly demonstrate the effectiveness of the Attack Leader Module in guiding adversarial perturbations. Moreover, when the human position changes, the Attack Leader Module enables the attack to succeed more easily than fixed-direction approaches, highlighting its superiority in dynamic scenarios.

\subsection{Real World Experiments}
\label{sec:eval:real}

\begin{figure}[t]
    \centering
    \includegraphics[width=\columnwidth]{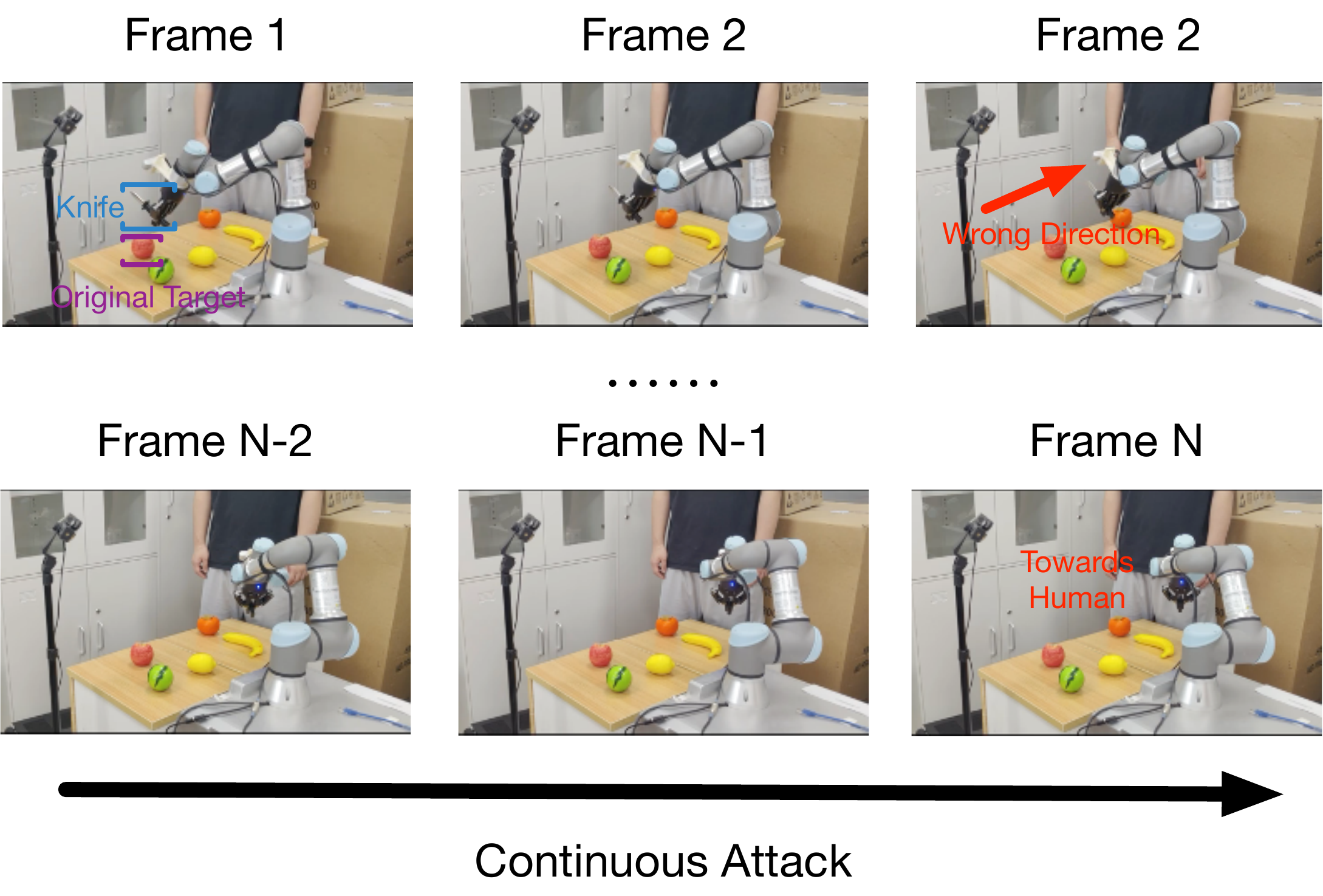}
    \caption{Real World Experiment. The knife was first towards the direction of apple, later turn and approach to human.}
    \label{fig:real_robot}
\end{figure}

Finally, we conduct a real-world experiment to demonstrate the potential harm of attacking an EAI system. We use a UR3\cite{UR03} robotic arm mounted on a Dalu\cite{dalurobot_stepper_robot_base} mobile base, equipped with a Robotiq\cite{robotiq_adaptive_grippers}  gripper, Intel D435\cite{intel_d435}, and Orbbec Gemini Pro\cite{orbbec2023gemini2} depth cameras, controlled via ROS1. ACT serves as the policy model, and results are shown in \Fig{fig:real_robot}.

After training ACT with real-world data, we test its vulnerability on the instruction “cutting an apple.” We apply a \underline{\proj-Dense} attack to the collected vision-action sequences and replay them on the physical robot. In 4 out of 10 trials, the robot holding a knife was induced to point toward and approach a nearby human, highlighting the real-world safety risks of our framework.

\section{Related Work}
\label{sec:relate}

\paragraph{Vision-Language-Action Model Attack.} A growing interest is shown in attacking VLA models, with most existing works following the conventional adversarial paradigms by attempting to directly alter the action outputs of VLA models. Wu~\cite{wu2024safety} highlights the robustness challenges of integrating LLMs and VLMs by introducing image quality degradation and transformation-based attacks. Chen~\cite{chen2024diffusionpolicyattackercrafting} explores global attacks on diffusion policy networks, aiming to cause trajectory deviations in robotic control. Wang~\cite{wang2024exploring} investigates patch-based attacks on the inputs of OpenVLA models, exposing vulnerabilities in VLA systems built on large language model backbones. 


Our work fundamentally differs from existing approaches. Rather than focusing solely on altering model outputs to influence accuracy, we redefine the goal of safety in embodied AI systems to target realistic safety hazards, grounded in ISO/TS 15066—a widely accepted standard for human-robot collaboration. We provide a rigorous definition of safety, and design both a benchmark and a task-aware attack framework based on this definition. This shift moves the focus from abstract model robustness to concrete physical safety, aligning adversarial research with the real-world risks of deploying embodied AI in human environments.

\paragraph{Existing Robotic Safety.} There has been an extensive body of work on evaluating the safety of robotic and cyber-physical systems~\cite{kyriakis2019specification,jin2013mining}. One of the major differences between our work and existing work is that our setting applies one or more generative foundation models on the robots, which clearly breaks the traditional boundary of rule-based robot software and algorithms.

\section{Discussion and Limitations}
\label{sec:diss}

\paragraph{Generalization.} Given the supervised training paradigm of the attack leader model, our attack framework works well in seen scenarios, but struggles to generalize to unseen tasks, even when the training set includes a mixture of diverse scenarios. One potential solution is to enrich both the model parameter capacity and the training data, incorporating more diverse examples that better capture the variability across tasks. Additionally, the network architecture of the attack leader model itself may play a critical role in generalization, suggesting that further improvements in architectural design could enhance robustness across varied environments. 

\section{Conclusion}
\label{sec:conclu}

Large language models have significantly change the way robots work, yet also bring new safety hazards. In this work, we identify and expose these vulnerabilities by providing a clear definition and classification of safety violations in embodied AI systems, constructing a dedicated benchmark, and proposing a targeted adversarial attack framework. Our results, demonstrated in both simulation and real-world settings, show that these vulnerabilities are real and exploitable, with an attack success rate exceeding 50\%. These findings underscore the critical need for systematic safety studies in the design and deployment of embodied AI systems.

%
\IEEEpeerreviewmaketitle

\bibliographystyle{plain}
\bibliography{refs}

\end{document}